\documentclass[runningheads]{llncs}
\usepackage{graphicx}
\usepackage{xcolor}
\usepackage[normalem]{ulem}

\begin{document}
\title{FeatGeNN: Improving Model Performance for Tabular Data with Correlation-based Feature Extraction\thanks{The author would like to acknowledge FAPEMIG (Fundação de Amparo à Pesquisa do Estado de Minas Gerais), CAPES (Coordenação de Aperfeiçoamento de Pessoal de Nível Superior), UFOP (Universidade Federal de Ouro Preto) and Cloudwalk, Inc, for the financial support which has been instrumental in the successful execution of our research endeavors.}}
\titlerunning{FeatGeNN}
%
\author{Sammuel Ramos Silva\inst{1,2}\orcidID{0000-0002-2274-1390} \and
Rodrigo Silva\inst{1}\orcidID{0000-0003-2547-3835}}
\authorrunning{S. Silva and R. Silva}

\institute{Universidade Federal de Ouro Preto, Ouro Preto 35402-163, Brasil\\
\email{sammuel.silva@aluno.ufop.edu.br}\\
\email{rodrigo.silva@ufop.edu.br}\\ \and
Cloudwalk, Inc, São Paulo, 05425-070 São Paulo, Brasil\\
\email{sammuel@cloudwalk.io}
}
\maketitle              
\begin{abstract}
Automated Feature Engineering (AutoFE) has become an important task for any machine learning project, as it can help improve model performance and gain more information for statistical analysis. However, most current approaches for AutoFE rely on manual feature creation or use methods that can generate a large number of features, which can be computationally intensive and lead to overfitting. To address these challenges, we propose a novel convolutional method called FeatGeNN that extracts and creates new features using correlation as a pooling function. Unlike traditional pooling functions like max-pooling, correlation-based pooling considers the linear relationship between the features in the data matrix, making it more suitable for tabular data. We evaluate our method on various benchmark datasets and demonstrate that FeatGeNN outperforms existing AutoFE approaches regarding model performance. Our results suggest that correlation-based pooling can be a promising alternative to max-pooling for AutoFE in tabular data applications.

\end{abstract}

\section{Introduction}

Creating effective features is a crucial aspect of machine-learning projects. Essentially, it involves deriving new features from existing data to train a model or extract more information for statistical analysis. Discovering novel features from raw datasets is often the key to improving model performance \cite{afew}. 

Traditionally, feature creation is a manual process that heavily relies on an analyst's domain knowledge and programming skills. However, this approach can be limiting, as an analyst's intuition and expertise often influence the features created. To overcome these limitations, researchers have been exploring the field of Automated Feature Engineering (AutoFE). AutoFE aims to automate the feature creation process, enabling the discovery of more complex and effective features without relying solely on human input.

Automated feature engineering methods involve applying transformations to raw data to create new features. One commonly used technique is the expansion-reduction method \cite{dfs}, which generates a large number of features and then applies a feature selection algorithm to reduce their dimensionality. During the expansion phase, various transformations, such as logarithmic, max/min, or sum, are applied to the raw data. In the reduction phase, a feature selection method is utilized to identify the most effective set of features, which can significantly enhance a model's performance.

The possible number of transformation operations that can be performed on already-transformed features is practically infinite, which leads to an exponential increase in the feature space. This issue can cause a problem in reducing the number of feature evaluations required. To address this issue, researchers have proposed adaptive methods for AutoFE. For instance, Khurana et al. \cite{khurama} introduced a Q-learning agent capable of performing feature transformation search, achieving higher performance but still generating a large number of features. In another study \cite{lfe}, a Multi-Layer Perceptron (MLP) was trained to suggest the best transformations for each raw feature, resolving the problem of excessive feature generation. More recently, DIFER \cite{difer}, a gradient-based method for differentiable AutoFE, has demonstrated superior performance and computational efficiency compared to other approaches, although it still requires significant computation

In recent years, the use of deep neural networks (DNNs) has become increasingly widespread across a range of fields, such as computer vision and natural language processing \cite{sq,nl}. Typically, these models extract new features by feeding input features into the hidden layers of a DNN. While this approach is effective in capturing complex interactions between implicit and explicit features, it may not always generate useful new features due to a lack of relevant interactions in the dataset \cite{ctr}. Moreover, most existing works use max-pooling in the pooling layer, which may not be optimal for tabular data because it does not preserve the order and context of features in the data matrix. Additionally, max-pooling is intended to identify the most significant features within an image, which may not always be relevant or effective for tabular data.

To address the limitations of existing AutoFE methods, we propose FeatGeNN, a convolutional approach that leverages correlation as a pooling function to extract and generate new features. FeatGeNN first applies convolutional filters to the raw data to extract high-level representations. Then, instead of using traditional pooling functions like max or average pooling, it computes the correlation between the extracted features, which helps to identify the most informative features. The selected features are then passed through a multi-layer perceptron (MLP) to create the final set of new features. Preliminary results indicate that FeatGeNN outperforms existing AutoFE methods in both the number of generated features and model performance, demonstrating its potential as a potent tool for creating features in machine learning.

\section{Related work}
The main goal of feature engineering is to transform raw data into new features that can better express the problem to be solved. Training a model with the generated features can increase the performance of the model. However, the process of feature engineering can be limited by the expertise, programming skills, and intuition of the person working with the data. For this reason, AutoFE approaches have recently gained attention.

The authors of \cite{difer} propose a differentiable AutoML model that efficiently extracts low and high-order features. The model includes three steps: Initialization, Optimizer Training, and Feature Evaluation. In initialization, features are constructed randomly and evaluated using a machine-learning model in the validation set. In training the optimizer, a tree-like structure is created with an encoder, a predictor, and a decoder, called a parse tree. The encoder maps the post-order traversal string to a continuous space, the predictor is a 5-layer MLP that maps the representation to the score computed by a machine learning model, and the decoder maps the embedding to the discrete feature space. In the final step of feature evolution, the best \textit{n} features are selected and optimized using a gradient-based approach.

In \cite{dfs}, the authors present an algorithm that uses mathematical functions to generate new features for relational databases. The algorithm begins by identifying the entities that make up the database and defines a set of mathematical functions that are applied at both the entity level and the relational level. The proposed approach first enumerates all possible transformations on all features and then directly selects features based on their impact on model performance. However, due to the potentially large number of features generated, it is necessary to perform feature selection and dimensionality reduction to avoid overfitting and improve the interpretability of the model.

In \cite{lfe}, the authors propose a novel model for feature engineering in classification tasks that can generalize the effects of different feature transformations across multiple datasets. The model uses an MLP for each transformation to predict whether it can produce more useful features than the original set. The Quantile Sketch Array (QSA) achieves a fixed-size representation of feature values to handle features and data of different lengths. The QSA uses Quantile Data Sketch to represent feature values associated with a class label.

The authors of \cite{nfs} have proposed an RNN-based approach to address the feature explosion problem in feature engineering and support higher-order transformations. Their architecture uses an RNN to generate transformation rules with a maximum order for each raw feature within a fixed time limit. For datasets with multiple raw features, the authors use multiple RNNs as controllers to generate transformation rules for each feature. The transformed features are evaluated using a machine learning algorithm and the controller is trained using policy gradients. The model includes two special unary transformations: "delete" and "terminate", which remove a feature and terminate the current transformation, respectively, to determine the most appropriate transformation order.

In \cite{khurama} they propose a heuristic model for automating feature engineering in supervised learning problems. Their model is based on a tree structure, where the raw dataset is the root, each node is a transformed dataset, and the edges represent the transformation functions. The goal is to find the node with the highest score, reducing the feature construction problem to a search problem.

The authors present three exploration strategies to traverse the tree. The first is depth-first traversal," in which a random transformation is applied to the root and the algorithm then explores a branch until there is no further improvement. Then it chooses another node with the highest score and starts the process again. The second is the "Global Traversal", where a global search is performed to find the most promising node out of all the nodes explored so far. The third is "Balanced Traversal", in which the algorithm chooses either an exploration or exploitation strategy at each step based on a time or node budget. To handle the explosive growth of columns, feature selection is required as they grow. Cognito allows the selection of features after each transformation to clean up the dataset and ensure a manageable size. In addition, at the end of the model execution, the algorithm performs another feature selection for all columns in the dataset, including the newly created columns.

AutoFeat is a method presented in \cite{autofeat} that generates and selects non-linear input features from raw inputs. The method applies a series of transformations to the raw input and combines pairs of features in an alternating multi-step process to generate new features. However, this leads to an exponential increase in the size of the feature space, so a subsampling procedure is performed before computing new features. The authors have shown that two or three steps of the feature technique are usually sufficient to generate new features.

After feature engineering, the new dataset has a higher number of features than the original dataset. To reduce the dimensionality, the authors developed a feature selection procedure. First, they remove new features that are highly correlated with the original or simpler features. Then they apply a wrapper method with L1-regular linear models to select the most informative and non-redundant features from the dataset. In the end, only a few dozen features are retained and used after the feature creation and selection process.

In \cite{autolearn}, autolearn is proposed, a learning model based on regression between pairs of features and aimed at discovering patterns and their variations in the data. The method selects a small number of new features to achieve the desired performance. The proposed method consists of four phases: Pre-processing to reduce dimensionality, where the authors perform feature selection based on information gain (IG); Mining of correlated features to define and search for pairwise correlated features, where the distance correlation \cite{dstcorr} is calculated to determine if there is an interesting predictive relationship between a pair of features; Feature generation, where regularized regression algorithms are used to search for associations between features and generate new features; and Feature selection, where features that do not add new information to the dataset are discarded.

The authors of \cite{wiseANDdeep} have proposed a novel model that achieves both memorization and generalization by simultaneously training a linear model component and a neural network component. The model consists of two components: The Wide component, which is a generalized linear model of the form $yW^txb$, where $y$ denotes prediction, $x$ denotes features, $w$ denotes model parameters and $b$ denotes bias. The input features can be either raw or transformed, the most important transformation being the cross-product transformation; and the Deep component, which is a feed-forward neural network. For categorical features, an embedding is created, which is then added to the dataset and fed into the network.

The authors of \cite{deepFM} have proposed a model for predicting CTR that can handle interactions between low and high-order features by introducing a factorization-machine (FM) based neural network. The model consists of two parts: the FM component, which generates low-order features and can generate interactions between 1st and 2nd-order features with low computational cost, and the deep component, a feed-forward neural network that learns interactions between higher-order features. The input to the network is a high-dimensional vector of sparse data containing categorical and continuous variables as well as grouped fields.

The FGCNN is another approach proposed in CTR for prediction\cite{ctr}. This model consists of two components, namely the Feature Generation and the Deep Classifier. The Feature Generation component uses the mechanisms inherent in the Convolutional Neural Network (CNN) and the Multilayer Perceptron (MLP) to identify relevant local and global patterns in the data and generate new features. The Deep Classifier component then uses the extended feature space to learn and make predictions.

Our work introduces a CNN-based model with correlation-pooling for extracting high-order features and improving model performance. Unlike traditional pooling functions such as max-pooling, which focus on selecting the maximum value within a pooling region, correlation-pooling considers the linear relationships between features in the data matrix. It measures the correlation coefficient between the features and aggregates them based on their correlation values to capture the interdependencies and patterns in the data. By incorporating correlation-based pooling into the feature extraction process, FeatGeNN can effectively extract high-order features that reflect the underlying relationships among input variables. Our proposed method achieves competitive results on a range of problems, suggesting that correlation-based pooling is a promising technique for working with tabular data in neural networks.

\section{Proposed Approach}
In this section, we describe the proposed Feature Generation with Evolutionary Convolutional Neural Networks (FeatGeNN) model in detail. 

\subsection{Problem Formulation}

Given an dataset $D = \left \langle F, tg \right \rangle$, where $F = \left \{ f_1, f_2, ..., f_n  \right \}$ are the raw features and $tg$ the target vector. We denote as $L_{E}^{M}(D,tg)$ the performance of the machine learning model $M$ that is learned from $D$ and measured by the evaluation metric $E$ (e.g. accuracy). In addition, we transform a raw set of features $D$ into $D_{new}$ by applying a set of transformation functions $T = \left \{ t_1, t_2, ..., t_n  \right \}$.

Formally, the goal of the AutoFE is to search the optimal transformed feature set $D^{*}$ where $L_{E}^{M}(D^{*},tg)$ is maximized.

\subsection{FeatGeNN Model}

\begin{figure}[h]
  \centering
  \includegraphics[width=12cm]{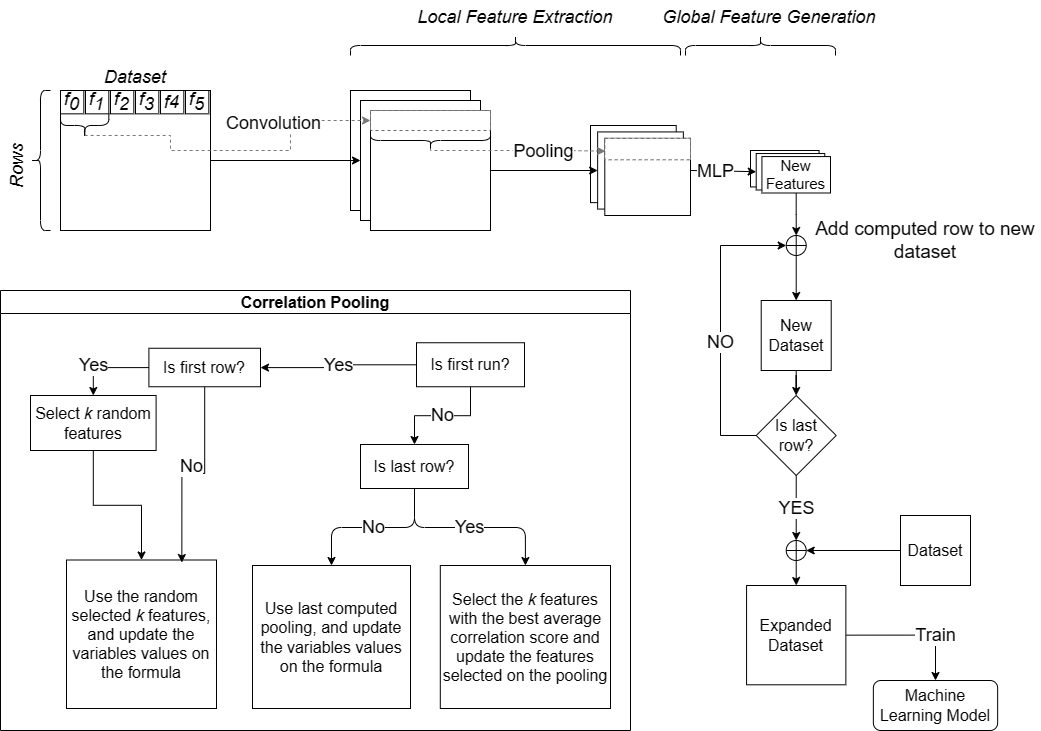}
  \caption{The FeatGeNN process.}
  \label{fig:model}
\end{figure}

In this study, we use a convolutional neural network to extract features that can improve the performance of a machine learning model (i.e., Random Forest). As explained earlier, using an MLP alone to generate new features would not result in a good set of new features. The reason for this is the relationship between the number of informative interactions between features and the total number of features in the feature space. Also, using a CNN alone might not lead to good performance because a CNN only considers local interactions and does not consider many important global interactions between features \cite{ctr}.

To overcome this problem, we use an architecture that combines the MLP with the CNN. The FeatGeNN model includes two main blocks, namely local feature extraction and global feature generation (Figure \ref{fig:model}). The first block attempts to identify the most informative interactions between local features, while the second block generates new features from the features extracted by the local feature extraction block and combines them globally.

The Local Feature Extraction block includes two main operations, namely Pooling and Convolution. Among these operations, the pooling operation plays a crucial role in reducing dimensionality and preserving the most informative features for subsequent layers. In previous work on feature generation for tabular data with CNN, max-pooling was mainly used. However, we found that using max-pooling for tabular data may not give the desired result because the model may not compare closely related features, thus affecting the features generated by the model. Therefore, we propose the use of correlation-pooling to address this issue.

\begin{figure}[h]
  \centering
  \includegraphics[width=12.5cm]{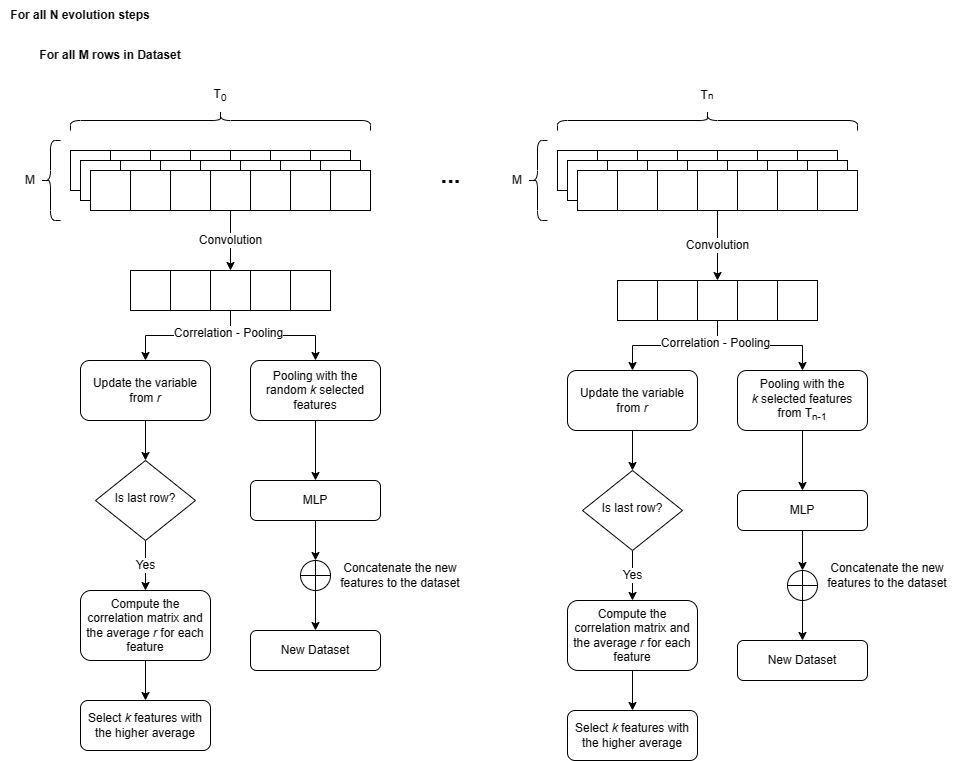}
  \caption{Correlation-Pooling process.}
  \label{fig:you
  correlation}
\end{figure}

In correlation pooling, the variant of pooling used in our Local Feature Extraction block, uses Pearson correlation\cite{pearson} to group features that are highly correlated. By grouping these features, correlation-pooling can preserve the relationship between closely related features and thus improve the quality of the features extracted by the CNN model. This is in contrast to max-pooling, which preserves only the most dominant feature in a group and may ignore other relevant features that are closely related. Therefore, by incorporating Pearson correlation in the pooling operation, correlation-pooling can effectively circumvent the limitation of max-pooling and help generate more informative features for subsequent layers in the CNN model. The Pearson correlation coefficient can be formulated as follows for our problem:

\begin{equation} 
r = \frac{n\sum xy - (\sum x)(\sum y)}{\sqrt{[n\sum x^{2}-(\sum x)^{2}][n\sum y^{2}-(\sum y)^{2}]}}
\end{equation}
where \textit{x} and \textit{y} represent the values of the features \textit{X} and \textit{Y} respectively, and $X,Y \in F$, where \textit{F} is the set of all features. The variable \textit{n} denotes the number of samples in the dataset \textit{D}. 

To avoid having to run the Pearson algorithm twice, we have introduced an iterative calculation of the Pearson coefficient. This means that at the current stage of model development, we compute the Pearson coefficient \textit{r} to perform the pooling operation for the subsequent evolutionary generation of the model. To reduce the computations required, we also added a threshold to limit the number of data sent to the correlation calculation, i.e., a model can only use 70\% of the data to calculate the correlation value for the features.

While the Pearson correlation is a statistical measure that describes the linear relationship between two variables, it is not suitable for analyzing relationships between more than two characteristics. To overcome this limitation, we use the multivariate correlation matrix, which consists of pairwise Pearson correlation coefficients between all pairs of variables. This matrix allows us to analyze relationships between multiple variables and identify the most highly correlated variables. The overall correlation value for the feature \textit{f} can be formulated as follows:

\begin{equation} 
CS_f = \frac{\sum_{k}^{N}r_{fk}}{N}
\end{equation}
where $CS_f$ is the correlation score for the feature \textit{f}, $r_{fk}$ represent the person correlation score for the feature tuple (\textit{f,k}) and \textit{N} the total number of feature in the dataset.

In the Global Feature Generation block, an MLP is utilized to merge the features extracted from the Local Feature Extraction block and generate novel features. These novel features are then appended to the original dataset and used in the machine-learning model.

\subsection{Evolution Process}

\begin{figure}[h]
  \centering
  \includegraphics[width=10.5cm]{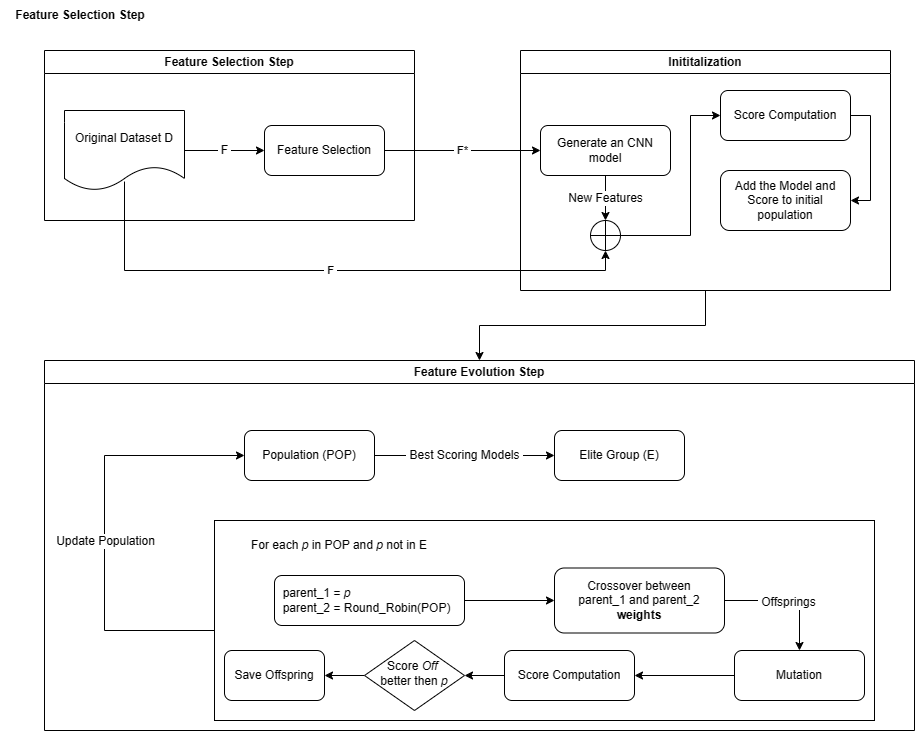}
  \caption{Feature Evolution process.}
  \label{fig:evolution}
\end{figure}

In this work, we adopt an evolution process for conducting AutoFE, as depicted in Figure \ref{fig:evolution}. This process involves three distinct steps: (1) Feature Selection, (2) Population Initialization, and (3) Feature Evolution.

The first step of our proposed approach is to reduce the combination of uncorrelated and redundant features using a feature selection method. We used the Maximum Relevance-Minimum Redundancy (MRMR) \cite{mrmr} method for this purpose. By minimizing the combination of such features, we aim to reduce the introduction of noise into the model and improve the quality of the features generated by the CNN model. 

During the population initialization step, we generate a population of the CNN model $POP$ that is evolved in the Feature Evolution step. To evaluate this initial population, we use a machine learning model on the dataset resulting from step (1). Specifically, we take the set of features \textit{F*} from the Feature Selection step and input them into the CNN model \textit{p} (where $p \in POP$) to generate \textit{n} new features \textit{f}. These newly created features are concatenated with the original dataset \textit{D} to create a new dataset $D* \left \{ F \cup f \right \}$, which is then evaluated by the machine learning model $L^m$ to obtain a score $S_p$.

In the trait evolution step, a genetic algorithm \cite{GA} is used to evolve the population and identify the most effective traits to improve the performance score obtained by $L^m$. During each epoch of the genetic algorithm, for each model $p$ that is not part of the elite group \textit{E} (where $E is \in POP$), a crossover is performed between its weights and those of a second model $p'$, which is selected using a round-robin tournament \cite{roundrobin}. Following the crossover process, the offspring generated by this operation can be subjected to mutation. The features produced by the offspring are then evaluated, as described in the initialization of the population initialization step. If the score obtained by $L^m$ is better than the current score for \textit{p} or if depreciation is allowed, the offspring replaces the current model \textit{p} and the score is updated.

\section{Results}
In this section, we aim to answer the following research questions:

\begin{itemize}
    \item \textbf{RQ1:} How effective is correlation-pooling compared to Max-Pooling?
    \item \textbf{RQ2:} Study of the impact of the number of data on the correlation pooling computation?
    \item \textbf{RQ3:} How effective is the proposed FeatGeNN approach? (Comparison with literature)
\end{itemize}

\subsection{Experimental Setup}

To evaluate the performance of the FeatGeNN model, on classification problems, 6 classification datasets from the UCI repository, which were used in the state-of-the-art methods \cite{difer} \cite{nfs}, were selected. The description of each dataset in terms of the Number of Features and Number of Samples is presented in Table \ref{tab:desc1}.

\begin{table}[h]
\centering
\caption{Statistics of the benchmarks used to perform the evaluation of the FeatGeNN features.}
\label{tab:desc1}
\begin{tabular}{l|cc}
\hline
Datasets    &  \multicolumn{1}{l}{Samples} & \multicolumn{1}{l}{Features} \\ \hline
SpamBase    &  4601                        & 57                           \\
Megawatt1   &  253                         & 37                           \\
Ionosphere  &  351                         & 34                           \\
SpectF      &  267                         & 44                           \\
Credit\_Default &  30000                    & 25                           \\
German Credit  & 1001                       & 24                           \\ \hline
\end{tabular}
\end{table}

In our experiments, we use the \textit{f1-score} as the evaluation measure, which is also commonly used in the related works \cite{nfs} and \cite{difer}. The threshold for questions RQ1 and RQ2 was set at 80\% of the available data in the dataset. To ensure robustness and reliability, we use 5-fold cross-validation, in which the dataset is divided into five subsets or folds and the evaluation is performed five times, with each fold serving once as a test set. This approach helps mitigate the effects of data variability and provides a more comprehensive assessment of the model's performance. As for the chosen algorithm, we use Random Forest as the base method in all our experiments. Random Forest is a popular and widely used ensemble learning method known for its robustness and ability to handle different types of data.

\subsection{Effectiveness of Correlation-Pooling vs. Max-Pooling (RQ1)}
In this subsection, this experiment aims to answer: \textit{Can our FeatGeNN with Correlation-Pooling achieve competitive results compared to the version with Max-Pooling?} Table \ref{tab:correlationxmax} shows the comparison results in terms of F1 score. The results show that the FeatGeNN with correlation-pooling outperforms the version with max-pooling in most datasets. The only exceptions are the Megawatt1 and Credit\_Default datasets, where the results are very similar. This result can be attributed to the fact that correlation-pooling takes into account the relationships between features when generating new features, which contributes to its relatively better performance.

\begin{table}[]
\centering
\caption{Comparing FeatGeNN performance with Correlation-Pooling and Max-Pooling. The * denotes the version of the FeatGeNN that was executed with Correlation-Pooling. The results are the average score, and the standard deviation, after 30 runs}
\begin{tabular}{l|ccc}
\hline
Dataset     & Base   & FeatGeNN       & FeatGeNN*      \\ \hline
SpamBase    & 0.9102 & 0.9422 (0.011) & \textbf{0.9530} (0.016) \\
Megawatt1   & 0.8890 & 0.9148 (0.002) & 0.9151 (0.002) \\
Ionosphere  & 0.9233 & 0.9587 (0.012) & \textbf{0.9667} (0.004)  \\
SpectF      & 0.7750 & 0.8682 (0.018) & \textbf{0.8776} (0.013) \\
Credit\_Default & 0.8037 &   0.8092 (0.003)  &   0.8095 (0.003)             \\
German Credit & 0.7401 &      0.7775 (0.006)          &     \textbf{0.7814} (0.002)            \\ \hline
\end{tabular}
\label{tab:correlationxmax}
\end{table}

\subsection{Impact of the Number of Data on the Correlation Pooling Computation (RQ2)}
\begin{figure}[h]
  \centering
  \includegraphics[width=12.5cm]{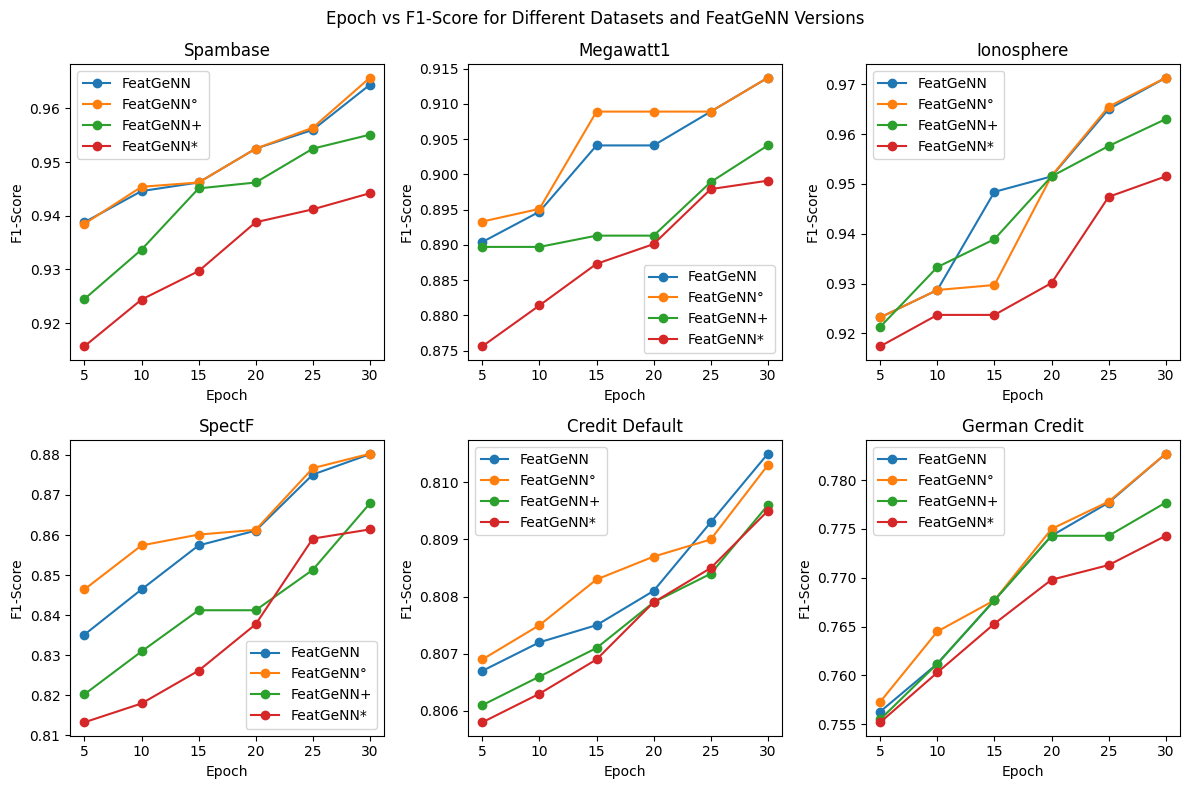}
  \caption{The performance of the different versions of FeatGeNN is compared in terms of the amount of data used for computation. In the image, the symbol represents the version that used 60\% of the available data, the * symbol represents the version that used 30\% of the data, and the symbol ° represents the version that used 100\% of the data. The FeatGeNN without symbol stands for the version that used 80\% of the available data.}
  \label{fig:difversions}
\end{figure}

In this subsection, our experiment aims to answer the question: \textit{What is the influence of the number of available data on the Correlation-Pooling computation?}. Figure \ref{fig:difversions} shows the performance of three versions of FeatGeNN: FeatGeNN (using all available data), FeatGeNN (using 60\% of the data), and FeatGeNN* (using 30\% of the data).

The results show that, as expected, the performance of the model varies with the amount of data used to compute the correlation-pooling. On average, the version with access to the entire dataset achieves a performance improvement of 0.76\% and 1.38\% compared to the FeatGeNN and FeatGeNN* versions, respectively. Compared to the version that used 80\% of the available data, the result after 30 epochs is very similar, although the version with more data performs better in fewer epochs. These results indicate that the performance of FeatGeNN is still competitive with the original version, even though the performance decreases slightly with less available data.

\subsection{Effectiveness of FeatGeNN (RQ3)}
In this subsection, this experiment aims to answer: \textit{Can our FeatGeNN with Correlation-Pooling achieve competitive results when compared to the state-of-the-art models?}. We compare FeatGeNN on 6 datasets with state-of-the-art methods, including (a) Base: Raw dataset without any transformation; (b) Random: randomly apply a transformation to each raw feature; (c) DFS \cite{dfs}; (d) AutoFeat \cite{autofeat}; (e) LFE \cite{lfe}; (f) NFS \cite{nfs}; and (g) DIFER \cite{difer}.

\begin{table}[]
\centering
\caption{Comparison between FeatGeNN with other methods from the literature, reported on \cite{difer}. * reports the average and standard deviation across 30 runs, while the FeatGeNN column reports the maximum value across the same runs.}
\begin{tabular}{l|cccccccc}
\hline
Dataset     & Base   & Random & DFS    & AutoFeat & NFS    & DIFER  & FeatGeNN*      & FeatGeNN \\ \hline
SpamBase    & 0.9102 & 0.9237 & 0.9102 & 0.9237   & 0.9296 & 0.9339 & 0.9530 (0.016) & \textbf{0.9644}   \\
Megawatt1   & 0.8890 & 0.8973 & 0.8773 & 0.8893   & 0.9130 & \textbf{0.9171} & 0.9151 (0.002) & \textbf{0.9171}   \\
Ionosphere  & 0.9233 & 0.9344 & 0.9175 & 0.9117   & 0.9516 & \textbf{0.9770} & 0.9644 (0.012)  & 0.9713   \\
SpectF      & 0.7750 & 0.8277 & 0.7906 & 0.8161   & 0.8501 & 0.8612 & 0.8776 (0.013) & \textbf{0.8802}   \\
Credit\_Default & 0.8037 & 0.8060 & 0.8059 & 0.8060   & 0.8049 & 0.8096 &   0.8095 (0.003)  & \textbf{ 0.8102}  \\
German Credit & 0.7410 & 0.7550 & 0.7490 & 0.7600   & 0.7818 & 0.7770 &   0.7814 (0.002)   & \textbf{0.7827}   \\ \hline
\end{tabular}
\label{tab:sota}
\end{table}

Table \ref{tab:sota} shows the comparative results of FeatGeNN relative to existing methods (results reported in \cite{difer}). From Table \ref{tab:sota} we can observe that in the classification tasks, the comparison shows that FeatGeNN, performs the best for the SpamBase, Credit\_Default, German Credit, and SpectF benchmarks, the second best for the Ionosphere benchmark and achieves the same result as the DIFER method for the Megawatt1 benchmark. Although DIFER achieves the best performance in the Ionosphere benchmark, they only achieve 0.58\% more than the best result obtained by our proposed method. 

Regarding the number of features, Table \ref{tab:featurescreated} shows that FeatGeNN excels in producing fewer features for the Megawatt1, SpectF, and Credit\_Default datasets compared to other methods. For the remaining datasets, FeatGeNN achieves comparable results with the same number of features. 

Compared to the performances of Base and Random, FeatGeNN achieved an average improvement of 5.89\% considering all datasets, which demonstrates the potential of the features generated by our proposed model.

\begin{table}[]
\centering
\caption{Comparison between FeatGeNN, DIFER, AutoFeat, and Random ($\ast$ the results reported on \cite{difer}).}
\begin{tabular}{lccccc}
\hline
Dataset         & Random & AutoFeat* & NFS* & DIFER* & FeatGeNN \\ \hline
SpamBase        & 1      & 46        & 57   & 1      & 1        \\
Megawatt1       & 8      & 48        & 37   & 29     & 8        \\
Ionosphere      & 1      & 52        & 34   & 1      & 1        \\
SpectF          & 8      & 37        & 44   & 9      & 8        \\
Credit\_Default & 4      & 30        & 25   & 5      & 4        \\
German Credit   & 1      & 22        & 24   & 1      & 1        \\ \hline
\end{tabular}
\label{tab:featurescreated}
\end{table}

\section{Conclusion}
In this study, we presented a novel approach for generating new features in tabular data that combines feature selection and feature generation to improve the performance of predictive models. Our proposed method uses a CNN architecture to effectively capture local features during convolution operations (Local Feature Extraction), thereby reducing the number of combinations required in the MLP phase (Global Feature Generation). In addition, we integrated a correlation-pooling operation as a dimensionality reduction step. Our approach demonstrates efficient feature learning and achieves competitive results compared to the architecture used by Max-Pooling and state-of-the-art methods.

As a direction for future research, we intend to explore information theory methods as possible alternatives for pooling operations. This could further increase the effectiveness of our approach to learning new features.

\end{document}